\documentclass[]{IEEEtran}
\usepackage{cite}
\usepackage{amsmath,amssymb,amsfonts}
\usepackage{algorithmic}
\usepackage{graphicx}
\usepackage{textcomp}
\usepackage{xcolor}
\usepackage{multirow}
\usepackage{caption} 
\usepackage{float}
\usepackage{soul}
\usepackage{xcolor}
\usepackage{amsmath}
\usepackage{hyperref}
\newcommand{\etal}{\textit{et al.}}

\usepackage{array}
\newcommand{\PreserveBackslash}[1]{\let\temp=\\#1\let\\=\temp}
\newcolumntype{C}[1]{>{\PreserveBackslash\centering}p{#1}}
\newcolumntype{R}[1]{>{\PreserveBackslash\raggedleft}p{#1}}
\newcolumntype{L}[1]{>{\PreserveBackslash\raggedright}p{#1}}
\usepackage{booktabs}

\usepackage{subcaption}
\hypersetup{
    colorlinks=true,
    allcolors=blue
}

\def\BibTeX{{\rm B\kern-.05em{\sc i\kern-.025em b}\kern-.08em
    T\kern-.1667em\lower.7ex\hbox{E}\kern-.125emX}}
\graphicspath{ {images/} }
\makeatletter
\let\old@ps@IEEEtitlepagestyle\ps@IEEEtitlepagestyle
\def\confheader#1{%
    \def\ps@IEEEtitlepagestyle{%
        \old@ps@IEEEtitlepagestyle%
        \def\@oddhead{\strut\hfill#1\hfill\strut}%
        \def\@evenhead{\strut\hfill#1\hfill\strut}%
    }%
    \ps@headings%
}
\makeatother

\begin{document}

\title{Huruf: An Application for Arabic Handwritten Character Recognition Using Deep Learning} 

\author{ 
    \IEEEauthorblockN{Minhaz Kamal\textsuperscript{*}, Fairuz Shaiara\textsuperscript{*}, Chowdhury Mohammad Abdullah\textsuperscript{*}, \\
    Sabbir Ahmed, Tasnim Ahmed, and Md. Hasanul Kabir\\}

    \IEEEauthorblockA{Department of Computer Science and Engineering,\\Islamic University of Technology, Gazipur, Bangladesh\\}
    
    \IEEEauthorblockA{Email: \{minhazkamal, fairuzshaiara, abdullah39, sabbirahmed, tasnimahmed, hasanul\}@iut-dhaka.edu}
}

\IEEEpubid{
\begin{minipage}[t]{\textwidth}\ \\[10pt]
      \small{* Contributed Equally\\
      (Copyright $\copyright$ 2022 IEEE. This work has been submitted to the IEEE for possible publication. Copyright may be transferred without notice, after which this version may no longer be accessible.)}
\end{minipage}
% \begin{minipage}[t]{0.5\textwidth}\ \\ [5pt]
%     \rule{0.4\textwidth}{0.1pt} \\
%      \small{* Contributed Equally}
% \end{minipage}
}

\maketitle

\begin{abstract}

Handwriting Recognition has been a field of great interest in the Artificial Intelligence domain. Due to its broad use cases in real life, research has been conducted widely on it. Prominent work has been done in this field focusing mainly on Latin characters. However, the domain of Arabic handwritten character recognition is still relatively unexplored. The inherent cursive nature of the Arabic characters and variations in writing styles across individuals makes the task even more challenging. We identified some probable reasons behind this and proposed a lightweight Convolutional Neural Network-based architecture for recognizing Arabic characters and digits. The proposed pipeline consists of a total of 18 layers containing four layers each for convolution, pooling, batch normalization,  dropout, and finally one Global average pooling and a Dense layer. Furthermore, we thoroughly investigated the different choices of hyperparameters such as the choice of the optimizer, kernel initializer, activation function, etc. Evaluating the proposed architecture on the publicly available `Arabic Handwritten Character Dataset (AHCD)' and `Modified Arabic handwritten digits Database (MadBase)' datasets, the proposed model respectively achieved an accuracy of 96.93\% and 99.35\% which is comparable to the state-of-the-art and makes it a suitable solution for real-life end-level applications. 
\end{abstract}

\section{Introduction}
Handwritten character recognition is a prominent aspect of computer vision with noticeable research efforts over the years. Handwriting recognition is of two types: online and offline. The offline characters are given as input through the scanned images from documents whereas the online characters are given input in run-time through the means of input device(s) \cite{online_offline}. This work deals with online characters. Handwriting recognition poses challenges of its own as can have very different handwriting \cite{ashikur2022twoDecades}. Nonetheless, an individual will have a slight variation in different characters each time they write \cite{PORWAL2013443}. It necessarily indicates that the challenge of handwritten character recognition revolves around the variability of pattern and distortion. This means feature extraction is the only way around it. Manual feature extraction does not help in such problems due to insufficient capturing of information.

Different algorithms, for example, Support Vector Machines (SVMs), Hidden Markov Models (HMMs), K-nearest neighbors (KNN), Neural Networks (NN), and Convolutional Neural Networks (CNNs) have been used in the literature to solve this problem statement at hand and most of them are for Latin languages \cite{7814123, 8281820, app9153169, ramzan2018survey, ahmed2019ocrBangla}. Among them, CNN turned out to be a better success. 

As a language of nearly 300 million people around the globe and also bearing religious significance for 1.8 billion Muslims around the world, Arabic also demands a well amount of research effort. Its characters are right to left, cursive, and also uses diacritical notations for vowels making the writing unique from other languages.
This is why a new learner from another language, for example, adults trying to learn the writing of Arabic characters needs to practice the writing for developing proper muscle memory. The same is applicable to children learning to write as well. Since the current population of all ages is exposed to digital devices like smartphones and personal computers, it can be used as a medium for easier access to learning and self-development. % \COR{Consequently, we worked on the idea of developing a web application for Arabic handwritten character recognition (AHCR).}

In this work, we present a recent review of literature in AHCR to understand the state-of-the-art works. Consequently, a comprehensive description of the CNN model followed by the result analysis is provided. A comparative analysis is presented to highlight the credibility of the model. Finally, the limitations and their potential solution is given before concluding the discussion.

\begin{figure*}[t]
    \centering
    \includegraphics[width=1\textwidth]{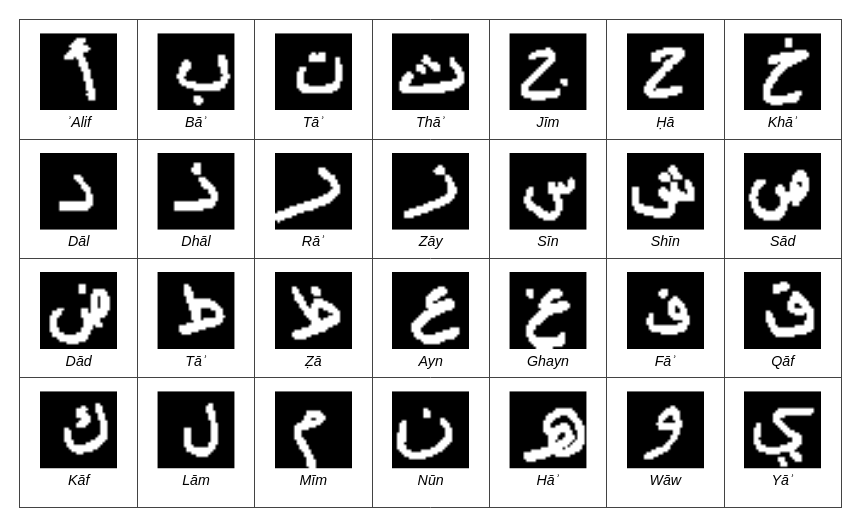}
    \caption{Samples from Each class of the AHCD dataset}
    % \captionsetup{justification=centering}
    \label{fig:example_char}
\end{figure*}

\section{Literature Review}
The first foundation of this domain came from the works of Latin character recognition. This is why starting with the analysis of this literature was important. Y. LeCun Was one of the pioneers to lead the way in handwritten character recognition. The works proposed by him relied on the core concepts of Neural Networks \cite{matan1990handwritten, lecun1995comparison}.  Classical machine learning algorithms were also used to approach this problem considering the easier implementation of the model and limited resources to train the models. In continuation, Y. LeCun himself published a ranking to compare their success \cite{firstRanking}.

From \cite{firstRanking}, it was seen that CNN-based solutions outmatch other techniques by far. But there were some classical machine learning solutions that presented fairly competitive accuracy. Using SVM \cite{decoste2002training} got the error rate between 0.56 to 0.68\%. Reference \cite{993558} and \cite{4250467} got the error rate between 0.56 to 0.68\% using KNN. Non-convolutional neural networks were also used by different authors and performed decently. But their reproducibility was criticized by \cite{martin_2016}. 

Finally, interesting outcomes were seen in the works based on Convolutional Neural Networks. From the author of \cite{lecun1995comparison}, a combination of different CNN architecture and adding data augmentation produced an error rate between  0.7 to 0.95\%. However, the most interesting work was seen from \cite{pmlr-v28-wan13} where they used DropConnect and their observed error rate was 0.21\% and 0.57\% with and without data augmentation.

Consequently, different kinds of literature were reviewed on the Arabic language. The first work published dates back to 2010 using Multi-Layer Perception (MLP) getting an accuracy of 94.3\% for digit recognition \cite{das2010handwritten}. The insights gathered for the Latin language suggested searching for CNN-based works since this approach has become state-of-the-art in this benchmark Pattern Recognition problem.  
A CNN-based model was proposed by \cite{younis2017arabic} for character recognition with an accuracy of 94.8\% and 94.6\% respectively in AIK9k and AHCD datasets. In a more recent paper, they used optimization, batch-normalization \cite{ioffe2015batch}, and regularization for better results and accelerated learning. A more recent paper \cite{ashiquzzaman2019efficient} used batch normalization and dropout in its CNN-based architecture. Using this model for the CMATERDB dataset, they got an astounding accuracy of 99.4\%. 

A recent work caught our attention which was not about AHCR but rather about Devanagari Character Recognition \cite{prashanth2022handwritten}. The reason behind it is that Devanagari characters are more cursive and complex in patterns compared to Latin characters. The paper also compared Modified versions of LeNet and Alexnet where Alexnet gave better accuracy.
% Finally, we searched for pre-trained models for the back-end of our application and found one from a github repository \cite{amrhendyrepo}. The model given incorporated almost all the noticeable aspects of the aforementioned works. The detailed description of the architecture is given in \ref{subsec:proposedMethod}.

\section{Methodology}

\subsection{Data Acquisition}
The dataset used for this project has been taken from the Kaggle kernels which include the Arabic Digits\footnote{Arabic Digits Dataset: \color{blue}\url{https://www.kaggle.com/datasets/mloey1/ahdd1}} and 
Arabic characters\footnote{Arabic Letters Dataset: \color{blue}\url{https://www.kaggle.com/mloey1/ahcd1}}. 
All the datasets were already converted to CSV files which contained the image pixel values and their respective labels. A set of examples is given in Figure~\ref{fig:example_char}.

\begin{figure}[ht]
    \centering
    \includegraphics[width=8.5cm]{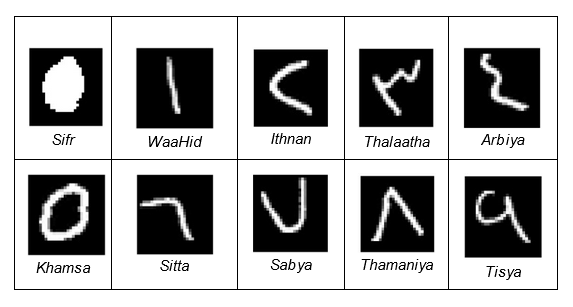}
    \caption{Samples form Each class of the MaDBase Dataset}
    % \captionsetup{justification=centering}
    \label{fig:example_digit}
\end{figure}

The Arabic Digits used are taken from a public dataset named `MaDBase'\footnote{MaDBase:  \color{blue}\url{http://datacenter.aucegypt.edu/shazeem/}} (Modified Arabic handwritten digits Database). It contains samples from 700 writers coming from separate institutions in order to capture the different writing styles. Each writer wrote 0 to 9 in Arabic ten times thus producing 70,000 images. A set of examples is given in Figure~\ref{fig:example_digit}.

The Arabic Letters have been taken from another public dataset names `Arabic Handwritten Character Dataset (AHCD)'. The dataset consists of 16,800 characters written by 60 participants where the ages ranged from 19 to 40. Each participant wrote `alef' to `yeh' ten times. This dataset considers only the first 28 characters of Arabic excluding `hamza' and thus produces 16,800 characters. The dataset is then further split into a training set of 13,440 characters which is further broken down into 28 classes having 480 images per class and a testing set of 3,360 characters which is again further divided into 28 classes each containing 120 images. The participants of the training and test set were completely different and the participants of the test set were randomized in order to avoid any kind of bias in producing the dataset.

The MaDBase and AHCD are the two largest databases \cite{de2018convolutional} available for this work and hence they are the most suitable datasets for training our model and achieving as accurate results as possible.

\subsection{Data Preprocessing}

\begin{figure}[b]
\centering
\begin{subfigure}{0.21\textwidth}
    \includegraphics[width=\textwidth]{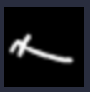}
    \caption{Initial Orientation}
    \label{fig:intitialalif}
\end{subfigure}
\hfill
\begin{subfigure}{0.21\textwidth}
    \includegraphics[width=\textwidth]{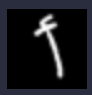}
    \caption{Changed Orientation}
    \label{fig:rotatedalif}
\end{subfigure}
\caption{The character `alef' from AHCD Dataset}
\label{fig:alifpreprocessing}
\end{figure}

As mentioned earlier, the datasets were already converted into CSV files and hence no further conversions were needed. Every pixel of the input images was re-scaled by dividing it by 255 in order to bring them in a range of [0,1]. Adding to that, the labels are all categorical values for multi-class classification problem. The output of the models can be labeled into the following categorical values:
\begin{itemize}
    \item Digits from 0 to 9 having categorical values from 0 to 9.
    \item Letters `alef' to `yeh' having categorical values from 10 to 37.
\end{itemize}
However, one key thing to mention is that the digits and the character dataset has been trained separately and hence the actual categorical values of the letters from `alef' to `yeh' will be 0 to 27.
One Hot Encoding technique with Keras has been used to encode these categorical values which essentially transformed the integers into a binary matrix consisting of only 1s and 0s.

TensorFlowJS has been used to link the model with an application in the backend. Thus Keras CNNs, when used with TensorFlow, require a 4D-array as input where the dimensions represent the number of sample images, the number of rows, the number of columns, and the number of channels of each image respectively. The 4D-arrays are also known as 4D tensors. The input images are of size $64\times64$ pixels and hence they need to be reshaped into 4D tensors of size (nb\_samples, 64, 64, 1). 

The images in the original dataset were present in a reflected manner and hence an adjustment had to be made before feeding the data to the input layer. The input was flipped and then rotated to resemble the actual image in the dataset before sending it to the input layer of the model. To clarify further the image of the letter `alef' in the original dataset has been shown in Figure~\ref{fig:intitialalif} and the image for the input data is shown in Figure~\ref{fig:rotatedalif}. 

\subsection{Proposed Method}
\label{subsec:proposedMethod}
After thoroughly investigating the literature, we observe that the recent Convolutional Neural Networks (CNNs) have been providing state-of-the-art performance in several classification tasks \cite{ahmed2019ocrBangla, morshed2022fruit, ahmed2022lessIsMore, yasmeen2021csvcNet}. Hence, we proposed a 18-layer CNN-based pipeline for this work. This section discusses the proposed model for this work elaborately.

\subsubsection{Input Layer}
The input images are all grayscale images having three dimensions that are height, width, and depth. An RGB image would have had three separate channels for red, green, and blue respectively but since the input images are grayscale, there will be a single channel instead of three. In this work, the model has as an input dimension of $64\times64\times1$ grayscale image.

In the input layer, at first 16 filters of size $3\times3$ have been used and Rectified Linear Unit (RELU) has been applied as the activation function in the input convolutional layer. Following the convolutional layer, a batch normalization layer has been applied to increase the training speed and to allow higher learning rates which makes the learning process easier as well as faster. Following that a max pooling layer and a 20\% random dropout layer are added. The max pooling layer reduces the number of features and selects the most dominant features which essentially increases the training speed and makes the features more robust. On the other hand, the dropout layer serves to alleviate the overfitting issue and thus gives a better generalization of the handwritten characters.

\begin{figure*}[ht]
    \centering
    \includegraphics[width=15cm]{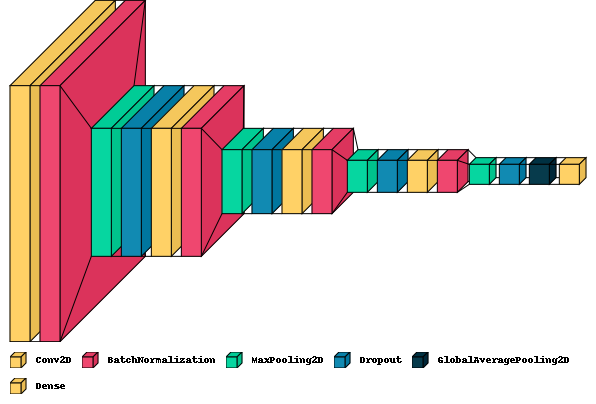}
    \caption{Visual representation of the CNN model for both digits and letters}
    % \captionsetup{justification=centering}
    \label{fig:cnn_model}
\end{figure*}

\subsubsection{Hidden Layers}
There are a total of 3 hidden layers in the pipeline. Each of the hidden layers consists of one convolutional layer, one batch normalization layer, one max pooling layer, and one random 20\% dropout layer. The three convolutional layers use 34, 64, and 128 number of filters of size $3\times3$ respectively. The hidden layers capture more details and complex patterns of the image such as corners, edges, endpoints, etc.

At the end, an additional layer of global average pooling has been applied. It is a replacement for the flattening function in Keras and it is more advantageous to use global average pooling instead of flattening because it closely resembles the convolutional structure and hence enforces the strong similarities between the filters and the categories.
In all the hidden layers, RELU has been used as the activation function as it gave the best results with the highest accuracy and minimum loss.

Additionally, categorical cross-entropy has been used as the loss function as the project deals with multi-class classification problems and accuracy has been used as a metric to achieve better performance from the model.

\subsubsection{Output Layer}
So far up to the hidden layer, the structure of the CNN model has been completely the same for both the digits as well as letters. However, in the output layer, the number of output classes will be different for both of them. The dataset consists of 10 digits from 0 to 9 and 28 letters from `alef' to `yeh'. Hence for the digit CNN model, there will be 10 neurons or output classes and for the letters CNN model there will be 28 neurons in the final layer. Both the models have been trained separately as stated earlier. The output layer uses the softmax activation function as it is the ideal one to use when dealing with multi-class classification problems where each output class will give the probability for that particular class.
A visual representation of the model can be seen in Figure~\ref{fig:cnn_model}. %Fig \ref{fig:cnn_model} has been generated using the Visualkeras package.

\subsubsection{Optimization}
For this work, a predefined model has been used with a total of 24 possible combinations of hyperparameter- which are optimizers, kernel\_initializer, and activation function. The different values that each hyperparameter are as follows:
\begin{itemize}
    \item optimizers: [`Adam', `RMSprop', `Nadam', `Adagrad']
    \item kernel\_initializer = [`uniform', `normal']
    \item activation = [`relu', `tanh', `linear']
\end{itemize}

A total of 5 epochs have been run for each combination of parameters. After comparing the results for all combinations the following combination has yielded the best result with maximum accuracy.
\begin{itemize}
    \item Optimizer: Adam
    \item Kernel\_initializer: uniform
    \item Activation: relu 
\end{itemize}

\section{Result Analysis}

\subsection{Experimental Setup}
The development and train-test experiment of the two models were done in the standard version of Google Colab. They were trained using the Adam optimizer, with a verbose of 1. We had 20 epochs with a batch size of 20. The training time for the Digit recognition model was comparably higher considering the large training set compared to Character recognition. The Digit recognition model took almost 45 minutes to complete whereas the Character recognition model took around 25 minutes to complete. Both of the models were saved in a Github repository.
Using tensorflow.js the model was used in a web application. ExpressJs framework with REST API was used for back-end development. EJS templating language with plain HTML, CSS, and JavaScript was used for front-end development\footnote{Implementation Details (Source Code): \url{https://github.com/minhazkamal/Arabic-Handwritten-Character-and-Digit-Recognition}}. 

\subsection{Performance Analysis}

\begin{table}[t]
\centering
\caption{Class-wise analysis on AHCD dataset}
\begin{tabular}{L{1.8cm} L{1cm} ccc}
\toprule
\textbf{Class Number} & \textbf{Class Label}  & \textbf{Precision} & \textbf{Recall} & \textbf{F1-score} \\ \midrule
0 & Alif                  & 0.98               & 0.99            & 0.99              \\
1 & Ba                    & 0.98               & 1.00            & 0.99              \\
3 & Ta                    & 0.91               & 0.96            & 0.93              \\
4 & Tha                   & 0.95               & 0.97            & 0.96              \\
5 & Jim                   & 0.99               & 0.99            & 0.99              \\
6 & Ha                    & 0.99               & 0.99            & 0.99              \\
7 & Kha                   & 0.97               & 0.99            & 0.98              \\
8 & Dal                   & 0.94               & 0.99            & 0.97              \\
9 & Dhal                  & 0.89               & 0.97            & 0.93              \\
10 & Ra                    & 0.94               & 0.99            & 0.97              \\
11 & Za                    & 1.00               & 0.88            & 0.93              \\
12 & Sin                   & 0.99               & 0.99            & 0.99              \\
13 & Shin                  & 0.99               & 1.00            & 1.00              \\
14 & Sad                   & 0.95               & 0.99            & 0.97              \\
15 & Dad                   & 1.00               & 0.93            & 0.97              \\
16 & Ta                    & 0.94               & 0.99            & 0.97              \\
17 & Zoa                   & 1.00               & 0.95            & 0.97              \\
18 & Aiyn                  & 0.96               & 0.98            & 0.97              \\
19 & Ghayn                 & 1.00               & 0.95            & 0.97              \\
20 & Fa                    & 0.90               & 1.00            & 0.95              \\
21 & Qaf                   & 0.99               & 0.91            & 0.95              \\
22 & Kaf                   & 1.00               & 0.97            & 0.99              \\
23 & Lam                   & 1.00               & 0.99            & 1.00              \\
24 & Mim                   & 0.98               & 0.98            & 0.98              \\
25 & Nun                   & 0.96               & 0.90            & 0.93              \\
26 & Ha                    & 0.98               & 0.95            & 0.97              \\
27 & Waw                   & 0.95               & 0.96            & 0.95              \\
28 & Ya                   & 1.00               & 0.96            & 0.98              \\ \midrule
\textbf{Accuracy}     &      &     &                 & 0.97              \\
\textbf{Macro Avg}   &  & 0.97              & 0.97            & 0.97              \\
\textbf{Weighted Avg}  & & 0.97               & 0.97            & 0.97 \\
\bottomrule

\end{tabular}

\label{tab:params_char}
\end{table}

\begin{table}[]
\centering
\caption{Class-wise analysis on the MaDBase dataset}
\begin{tabular}{llccc}

\toprule
\textbf{Class Number}   &   \textbf{Class Label}  & \textbf{Precision} & \textbf{Recall} & \textbf{F1-score} \\ 
\midrule
0   &   Sifr                 & 0.99               & 0.99            & 0.99              \\
1   &   Wahid                 & 0.99               & 1.00            & 0.99              \\
2   &   Ithnan                & 0.98               & 1.00            & 0.99              \\
3   &   Thalaatha             & 1.00               & 0.99            & 1.00              \\
4   &   Arbiya                & 1.00               & 0.99            & 0.99              \\
5   &   Khamsa                & 0.99               & 0.99            & 0.99              \\
6   &   Sitta                 & 0.99               & 1.00            & 1.00              \\
7   &   Sabya                 & 1.00               & 1.00            & 1.00              \\
8   &   Thamaniiya            & 1.00               & 0.99            & 1.00              \\
9   &   Tisya                 & 1.00               & 0.99            & 1.00              \\ \midrule
\textbf{accuracy}     &    &                &                 & 0.99              \\
\textbf{macro avg}    &  & 0.99               & 0.99            & 0.99              \\
\textbf{weighted avg} &  & 0.99               & 0.99            & 0.99   \\
\bottomrule
\end{tabular}

\label{tab:params_digit}
\end{table}

The final result for the characters showed a test accuracy of 96.93\%. On the other hand, we achieved an accuracy of 99.35\% for handwritten digit recognition.
%The Accuracy vs the number of epochs visualization can be seen in fig. \ref{fig:accuracyVSepoch}. 
A quantitative analysis is also needed to evaluate a model. The following parameters and their definitions are therefore relevant.

\begin{itemize}
  \item Recall (R): The parameter denoting the fraction of rightly classified sample over the total number which is from a particular class x. The expression will be : $$Recall=\frac{TP}{TP+FN}$$ 
  where $TP=True Positive, FN=False Negative$. 
  \item Precision (P): The parameter denoting the fraction of samples that are rightly classified over the total number of samples. The expression is: $$Precision=\frac{TP}{TP+FP}$$
  where $FP=False Positive$.
  \item F1 Measure: It combines the previous two parameters. The expression is: $$F1-score=\frac{2\times P\times R}{P+R}$$
\end{itemize}

The class-wise analysis for Arabic handwritten characters and digit recognition are respectively available in Table~\ref{tab:params_char} and Table~\ref{tab:params_digit}.

% The test accuracy of the proposed model was also compared against a vanilla CNN model and a model with the classical Alexnet architecture. Table \ref{tab:comparingarchi} shows the comparison of accuracy for these three architectures. Here we can see that the vanilla CNN model performs very poorly and should be discarded from the discussion. For the AlexNet, we can see the accuracy is not satisfactory and a higher loss is there. We think it is due to the fact that we considered parameter tuning in our model. We did not try LeNet because it is already proven that LeNet will not give better performance \cite{prashanth2022handwritten}. 

% \begin{table}[tb]
% \centering
% \begin{tabular}{l cc cc }
% \toprule
% \multicolumn{1}{c }{\textbf{Architecture}} & \multicolumn{2}{c}{\textbf{Test Accuracy (\%)}} & \multicolumn{2}{c}{\textbf{Test Loss (\%)}} \\ \cline{2-5} 
% & \multicolumn{1}{c}{Letters} & Digits & \multicolumn{1}{c}{Letters} & Digits \\ \midrule
% \textbf{Vanilla CNN}    & \multicolumn{1}{c}{8.21}    & 32.03  & \multicolumn{1}{c}{308.61}  & 177.86 \\ \midrule
% \textbf{Proposed Model} & \multicolumn{1}{c}{96.93}   & 99.35  & \multicolumn{1}{c}{10.60}   & 2.45   \\ \midrule
% \textbf{Alexnet}        & \multicolumn{1}{c}{90.38}   & 98.79  & \multicolumn{1}{c}{37.94}   & 7.38   \\ \bottomrule
% \end{tabular}
% \caption{Comparison of architectures of CNN}
% \label{tab:comparingarchi}
% \end{table}

\begin{table}[h]
\centering
\caption{Performance Comparison with the existing works on AHCD dataset}
\begin{tabular}{l c}
\toprule
\multicolumn{1}{c}{\textbf{Reference}}   & \textbf{Test Accuracy (\%)} \\ \midrule
 
Djaghbellou \etal \cite{djaghbellou2020arabic}       & 74.61 \\ 
Younis \etal \cite{younis2017arabic}    & 94.70    \\ 
Alkhateeb \etal \cite{alkhateeb2020effective} & 95.41   \\
Huruf (ours)                      & 96.93\\                   
\bottomrule
\end{tabular}

\label{tab:compare_char}
\end{table}

\begin{table}[h]
\centering
\caption{Performance Comparison with the existing works on  MaDBase dataset}
\begin{tabular}{l c}
\toprule
\multicolumn{1}{c}{\textbf{Reference}}    & \textbf{Test Accuracy (\%)} \\ \midrule

Loey \etal \cite{loey2017deep}         & 98.51   \\     
Abdleazeem \etal \cite{abdleazeem2008arabic} & 98.81   \\ 
Sousa \etal \cite{de2018convolutional}  & 99.47 \\
Huruf (ours)                       & 99.35  \\
\bottomrule
\end{tabular}
\label{tab:compare_digit}
\end{table}

Going further with our model, we can compare the results with some of the existing research works also. Table~\ref{tab:compare_char} compares the performance of the proposed work with the existing works on the AHCD dataset. In the same way, a comparison for the MaDBase dataset is presented in Table~\ref{tab:compare_digit}. For both the AHCD and MaDBase datasets, our model is showing comparable results.

\section{Conclusion and Future Works}

A fast and accurate framework for character and digit recognition can go a long way in providing automation for applications requiring robust handwriting processing. In this regard, we have proposed a CNN-based solution for Arabic handwritten character recognition. The proposed pipeline consists of a number of convolution, pooling, batch normalization, etc., layers and has achieved comparable accuracy to the state-of-the-art.
However, the model still has room for improvement as some letters were not identified correctly even after writing them in a proper way (for example letters without any dots). This is due to the quality of the AHCD dataset. Since there are lots of madrasas in Bangladesh and a good number of people can write Arabic letters properly, hence, a larger dataset can be curated with a higher number of samples. The application does not have any feature of segmenting characters from continuous handwriting. Incorporating this can be an interesting area to be explored in the future. %Using more sophisticated classes of ImageNet requires more pre-processing of data since they work with RGB data only. So this is also left for future work.
% Our proposed model can be thought of as a new edition for the AHCR domain showing promising results. It can be used by researchers for developing their own model followed by appropriate data preprocessing. 
% Using transfer learning can be directly used in other similar languages. For example, it can be directly used in Urdu or Persian language due to the similarity of their writings. This will also require further modification and testing of the architecture. 
\bibliographystyle{ieeetr}
\bibliography{reference}
\end{document}